\pdfoutput=1
\documentclass{article}

\usepackage{graphicx}
\usepackage{amsmath}
\usepackage[numbers]{natbib} % or authoryear
\usepackage[hidelinks]{hyperref} % load after natbib

\begin{document}

\title{Power-based Partial Attention: Bridging Linear-Complexity and Full Attention}

\author{
    Yufeng Huang\thanks{Corresponding author: yufeng@concavity.ai} \\
    concavity.ai \\
    \texttt{yufeng@concavity.ai}
}
\date{}

\maketitle

\begin{abstract}
It is widely accepted from transformer research that "attention is all we need", but the amount of attention required has never been systematically quantified. 
Is quadratic $O(L^2)$ attention necessary, or is there a sub-quadratic attention mechanism that can achieve comparable performance? 
To answer this question, we introduce power-based partial attention (PPA), an attention mechanism of order $O(L^{1+p})$, where $0 \leq p \leq 1$, such that 
$p=0$ corresponds to sliding window attention with linear complexity, and $p=1$ corresponds to full attention. 
With this attention construction, we can explore how transformer architecture performance varies as a function of the attention scaling behavior controlled by $p$. The overall trend from our experiments
shows an S-curve-like behavior where the performance transitions from sliding-window (linear-complexity) attention to full attention over a narrow window of $p$ values, and plateaus as $p$ approaches $1$. 
In our experiments, we show that there exists $0<p<1$ such that $O(L^{1+p})$ attention is sufficient to achieve similar results as $O(L^2)$ full attention. 

\end{abstract}

\section{Introduction}
The transformer architecture's self-attention mechanism has become the foundation of modern large language models (LLMs), 
enabling them to capture long-range dependencies and contextual relationships that drive remarkable performance across diverse tasks. 
However, this power comes at a steep cost: the standard self-attention mechanism scales quadratically with sequence length, 
requiring $O(L^2)$ time and memory for a sequence of length $L$. 
As applications increasingly demand longer context windows—for tasks such as document understanding, code generation, and extended dialogues—this 
quadratic bottleneck has become critical. Doubling the sequence length quadruples both computational cost and memory requirements, 
making context windows beyond tens of thousands of tokens prohibitively expensive.

This has motivated extensive research into efficient attention mechanisms that reduce complexity while preserving performance. 
However, a fundamental question remains unanswered: \textit{how much attention is actually necessary?} 
Is the full $O(L^2)$ attention truly required, or can sub-quadratic mechanisms achieve comparable results? 
Understanding this trade-off is essential for designing efficient architectures, yet prior work lacks a systematic framework 
to directly quantify the relationship between attention complexity and model performance.

Existing work on efficient attention can be categorized by computational complexity:
\textbf{(1) $O(L^2)$ sparse attention} reduces computation through sparsity while maintaining quadratic scaling. 
Early approaches like Big Bird \cite{zaheer2021bigbirdtransformerslonger} used random sparse patterns. 
Recent advances include hardware-aligned methods like native sparse attention (NSA) \cite{yuan2025nativesparseattentionhardwarealigned, yan2025fsaalternativeefficientimplementation}, 
block-level approaches like Mixture of Block Attention (MoBA) \cite{lu2025mobamixtureblockattention, xiao2025optimizingmixtureblockattention}, 
and token-selection methods like DeepSeek sparse attention (DSA) \cite{deepseekai2025deepseekv32pushingfrontieropen}. 
While these reduce constant factors, they remain fundamentally $O(L^2)$.
\textbf{(2) $O(L\sqrt{L})$ methods} include Sparse Transformer \cite{child2019generatinglongsequencessparse} and Monarch Attention \cite{yaras2025monarchattentionzeroshotconversionfast}, 
though they typically target encoder-style rather than causal attention. 
The Routing Transformer \cite{roy2020efficientcontentbasedsparseattention} uses K-means clustering with $O(L\sqrt{L})$ complexity, 
performing similarly to the Reformer \cite{kitaev2020reformerefficienttransformer}.
\textbf{(3) $O(L\ln{L})$ log-linear attention} includes the Reformer's LSH clustering \cite{kitaev2020reformerefficienttransformer}, 
Log-Linear Attention using Fenwick trees \cite{guo2025loglinearattention}, and trainable Log-Linear Sparse Attention (LLSA) \cite{zhou2025trainableloglinearsparseattention}.
\textbf{(4) $O(L)$ linear-complexity attention} encompasses state space models and commonly termed linear attention variants, including 
Linformer \cite{wang2020linformerselfattentionlinearcomplexity}, RWKV \cite{peng2023rwkvreinventingrnnstransformer}, 
Gated Linear Attention Transformers \cite{yang2024gatedlinearattentiontransformers}, Delta Networks \cite{schlag2021lineartransformerssecretlyfast}, 
Gated Delta Networks \cite{yang2025gateddeltanetworksimproving}, and Mamba \cite{ali2024hiddenattentionmambamodels, dao2024transformersssmsgeneralizedmodels}. 
Sliding window \cite{beltagy2020longformerlongdocumenttransformer} and streaming attention \cite{xiao2024efficientstreaminglanguagemodels} also achieve linear complexity.
\textbf{(5) Hybrid architectures} mix linear and full attention layers to balance efficiency and performance, 
including MambaFormer \cite{park2024mambalearnlearncomparative}, Jamba \cite{lieber2024jambahybridtransformermambalanguage}, 
Nemotron-H \cite{nvidia2025nemotronhfamilyaccurateefficient}, Qwen3-Next (combining gated delta net \cite{yang2025gateddeltanetworksimproving} 
and gated attention \cite{qiu2025gatedattentionlargelanguage}), and Kimi-linear \cite{kimiteam2025kimilinearexpressiveefficient}. 

A critical gap exists between $O(L)$ sliding window attention and $O(L^2)$ full attention. 
Despite similar hybrid architectures, models using sliding window attention significantly underperform those with full attention, 
and this performance gap remains poorly understood. 
To address this, we introduce \textbf{power-based partial attention (PPA)}, a parameterized attention mechanism with complexity $O(L^{1+p})$ where $0 \leq p \leq 1$. 
This framework enables systematic exploration of the attention complexity-performance trade-off: $p=0$ corresponds to sliding window attention with linear complexity, 
while $p=1$ recovers full quadratic attention. By varying $p$, we can precisely characterize how model performance transitions between these extremes.
Recently and concurrent with our work, PowerAttention \cite{chen2025powerattention} explored a related power-based sub-quadratic attention direction. While both efforts share the high-level motivation, our focus here is a simple causal masking construction (incremental stride selection combined with sliding window attention) and a controlled sweep over $p$ to directly quantify the resulting complexity--performance transition.

Our key contributions are:
\begin{itemize}
\item We introduce a simple, parameterized causal attention-mask family with tunable complexity $O(L^{1+p})$ that continuously interpolates between sliding-window (linear-complexity) attention and full (quadratic) attention.
\item We provide a controlled empirical characterization of the complexity--performance relationship as a function of $p$, revealing an S-shaped transition with rapid improvement in a narrow range (around $p\approx0.75$--$0.875$ in our experiments).
\item We show that sub-quadratic attention ($p < 1$) can achieve near-full-attention performance on MATH500 and GSM8k after task-specific adaptation.
\item We find that the optimal $p$ depends on training data quantity and prompt structure, suggesting that additional training can lower the required attention complexity.
\end{itemize}

These findings provide both theoretical insight into attention's role in transformers and practical guidance for designing efficient architectures that balance performance and computational cost.

\section{Methodology}

\begin{figure}[htbp]
    \centering
    \includegraphics[width=\textwidth]{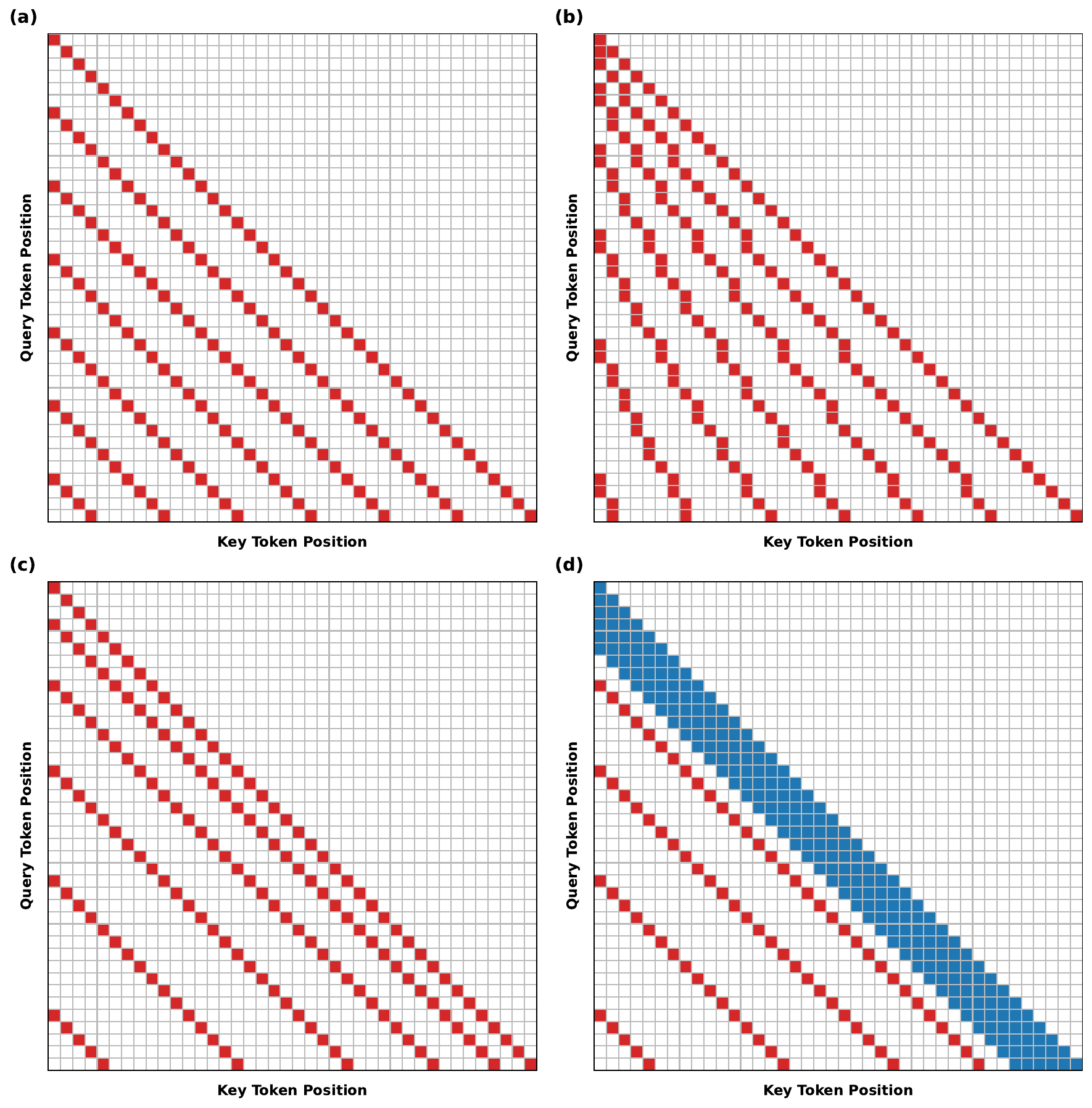}
    \caption{
    Illustrations of attention masks for (a) fixed stride attention, (b) dynamic stride attention, (c) incremental stride attention, 
    and (d) incremental stride combined with sliding window attention. 
    Each row represents a query token position, and each column represents a key token position. 
    In fixed stride attention, each token attends to previous tokens at fixed intervals (strides). 
    In dynamic stride attention, each token attends to previous tokens at intervals that scale with the sequence length. 
    In incremental stride attention (red cells), the stride increases incrementally for each token, 
    allowing for more comprehensive coverage of previous tokens while maintaining sub-quadratic complexity.
    In (d), blue cells show the sliding window attention overlay that ensures local context visibility, 
    which is combined with the incremental stride pattern (red cells where not overlapped by the sliding window) 
    to form the complete power-based partial attention mechanism used in our experiments.}
    \label{fig:stride_attention}
\end{figure}

Prior work \cite{child2019generatinglongsequencessparse} on dynamic stride attention proposes using a stride size of $\sqrt{L}$ rather than a fixed stride of size $k$.
While fixed stride sizes work for both encoder and decoder attention masks, the $\sqrt{L}$ dynamic stride attention can only be applied to encoders 
since the stride size depends on the global sequence length $L$. One could modify this approach for the decoder case by using a position-dependent mask 
where the $i$-th token attends to $\sqrt{i}$ tokens separated by intervals of $\sqrt{i}$. However, such a modified attention mask no longer 
has a regular strided pattern, making it difficult to optimize for efficient attention computation.

A strategy to overcome this issue is to use incrementally increasing stride lengths that preserve both the strided pattern and the $\sqrt{L}$
scaling for each token, so that we can perform $O(L^{\frac{3}{2}})$ attention on the decoder-only model. 
To construct such a strided pattern, we ensure that each token attends to at most $O(\sqrt{L})$ tokens. 
The main idea is to make each token $i$ to attend to $O(\sqrt{i}) \leq O(\sqrt{L})$ past tokens, so that the overall scaling for the whole 
sequence is $O(L^{\frac{3}{2}})$. 

We can view this as the density of attention, such that the average amount of attention per token is then $\frac{1}{\sqrt{i}}$ for token $i$. 
Then, we can integrate the density to calculate the total number of attended tokens as $\int_0^i \frac{1}{\sqrt{j}} dj = \frac{1}{2} j^{\frac{1}{2}}\big|_0^i = \frac{1}{2} i^{\frac{1}{2}} \approx O(\sqrt{i})$.
Performing another integration to sum over all tokens yields $\int_0^L i^{\frac{1}{2}} di = \frac{2}{3} L^{\frac{3}{2}} \approx O(L^{\frac{3}{2}})$ for the whole sequence. 

Using this strategy, the token positions to attend to can be taken from the squared values $1, 4, 9, 16, 25, 36, \ldots$ where $1$ is the current token position, 
$4$ is three positions back, and so on. This pattern preserves the $\sqrt{L}$ scaling for each token. For example, for a sequence of length $25$, 
the $25$th token attends to only $\sqrt{25} = 5$ tokens. We can extend this methodology to general $p=1/n$, where $n$ is a positive integer. 
For instance, $p=\frac{1}{3}$ corresponds to attending to positions $1, 8, 27, 64, 125, 216, \ldots$ (cubic positions). 
We can generalize this methodology for any $0<p<1$, such that whether position $j$ is an attended position can be determined
by checking if $\lfloor j^{p}\rfloor$ increments by 1, where $\lfloor \cdot \rfloor$ denotes the floor function.

For example, when $p=1/2$ and $j=9$, we have $\lfloor 8^{1/2}\rfloor = 2$ and $\lfloor 9^{1/2}\rfloor = 3$. Thus, the attention mask at position $j$ 
relative to the current token $i$ can be computed as $\lfloor j^p \rfloor - \lfloor (j-1)^p \rfloor$, which equals either $0$ or $1$. 
For the case of $j=9$, we have $\lfloor 9^{1/2} \rfloor - \lfloor 8^{1/2} \rfloor = 1$, indicating that $9$ is an attended position.
On the other hand, for $j=8$, we have $\lfloor 8^{1/2} \rfloor - \lfloor 7^{1/2} \rfloor = 0$, indicating that $8$ is not an attended position
when $p=1/2$. We can perform the same calculation for $p=1/3$, and one would find that $8$ is indeed the attended position in the case 
of $p=1/3$. 

It should be noted that the distance between consecutive attending positions grows sublinearly with sequence length, scaling like $O(L^{1-p})$. For example, when $p=1/2$, we can 
index the attending position by $j$, then the relative positions of the attending tokens can be obtained easily as the squared values of $j$, 
which means that index values of $j=1, 2, 3, 4, 5, 6, \ldots$ correspond to the attending positions of $1, 4, 9, 16, 25, 36, \ldots$ from earlier.
Using the index values, the distance can be calculated using binomial expansion as $j^{1/p}-(j-1)^{1/p} \approx \frac{1}{p} j^{1/p-1}$. Since $j \le L^p$, the maximum gap scales like $O(L^{1-p})$. 
For example, when $p=1/2$, the distance between two consecutive attending positions is $j^2 - (j-1)^2 = 2j - 1$. As a more concrete example, let the index $j=6$, then the attending position 
is $j^2 = 36$, and the distance from the previous attending position indexed by $j=5$ is $2\times j-1 = 2 \times 6 - 1 = 11$, which is just $36-25$.

\section{Implementation and Experimentation}

Using the power-based attention mechanism described above, we can vary $p$ to observe the transition from sliding window attention ($O(L)$) to full attention ($O(L^2)$).
To use this mechanism more effectively, we combine it with sliding window attention. Since sliding window attention has $O(L)$ complexity,
it does not affect the overall asymptotic scaling, but it is crucial for generating coherent text by making recent tokens visible to the model. By taking 
the union of the power-based attention mechanism and sliding window attention, and varying $p$ from $0$ to $1$, we can directly observe how the model transitions
from pure sliding window attention to full attention. 

For our implementation, we use the nvidia/NVIDIA-Nemotron-Nano-9B-v2 model \cite{nvidia2025nvidianemotronnano2}, a recent hybrid-attention model with a manageable size. 
We modify the attention layer so that the causal mask is the union of the power-based attention mechanism and sliding window attention. We introduce
the power parameter $p$ to control the attention scaling exponent. Since we are primarily interested in model output performance, we use PyTorch's built-in 
flex-attention \cite{dong2024flexattentionprogrammingmodel} for straightforward implementation. 

After modifying the attention layer, we evaluate the performance for each value of $p$ on the MATH500 benchmark \cite{hendrycks2021measuringmathematicalproblemsolving}. 
However, because the model was trained on full attention, simple fine-tuning is needed to adapt the model to this new architecture.
We chose the math subset of nvidia/Nemotron-Post-Training-Dataset-v2 \cite{nvidia2025nvidianemotronnano2} to fine-tune the model at different $p$ for this math task.  
To make the training more effective, we preprocess the dataset to have a similar 4-shot prompting format as the evaluation of MATH500 \cite{lewkowycz2022solvingquantitativereasoningproblems}.
After training, in addition to MATH500, we also evaluate each $p$ on GSM8k \cite{cobbe2021trainingverifierssolvemath}, which uses a different prompt format, 
this additional test allows us to observe how performance shifts when the models are insufficiently trained on that particular format. 

Finally, to measure raw performance without extended reasoning traces, we disable the thinking mode by using \texttt{<think></think>} in the generation prompt, rather than \texttt{<think>}. 

\section{Results and Discussion}

\begin{figure}[htbp]
    \centering
    \includegraphics[width=0.95\textwidth]{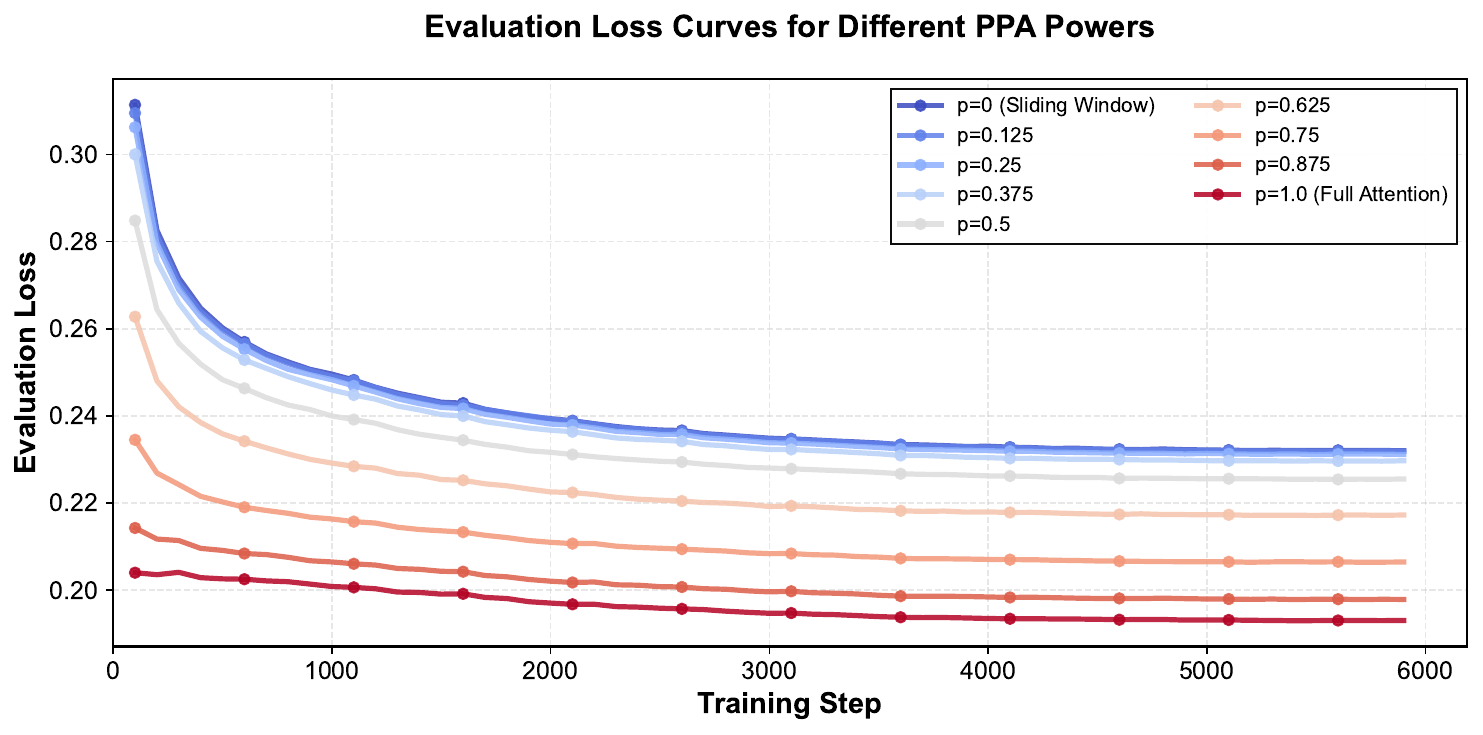}
    \caption{Evaluation loss curves during fine-tuning for different PPA power values. 
    The curves show a clear ordering by $p$, with higher $p$ consistently achieving lower evaluation loss. 
    The most pronounced separation occurs between $p=0.5$ and $p=0.875$, foreshadowing the performance hierarchy observed in the final benchmark results.}
    \label{fig:training-curves}
\end{figure}

We fine-tuned nine variants of the model, each with a different power parameter $p$ ranging from 0 (sliding window only) to 1 (full attention), 
using 200k examples from the math subset of nvidia/Nemotron-Post-Training-Dataset-v2. All models were trained with a sliding window size of 64 tokens 
and a 4-shot prompting format matching the MATH500 evaluation setup. Figure~\ref{fig:training-curves} shows the evaluation loss curves 
for all variants during fine-tuning.

The fine-tuning dynamics provide insight into how different PPA variants adapt to the task. 
The evaluation loss curves in Figure~\ref{fig:training-curves} show a clear ordering by $p$, with higher $p$ consistently achieving better generalization. 
The smooth convergence patterns suggest that the PPA mechanism can be fine-tuned stably across the full range of $p$, 
including intermediate values where the attention pattern is neither purely local nor fully global.

\begin{figure}[htbp]
    \centering
    \includegraphics[width=\textwidth]{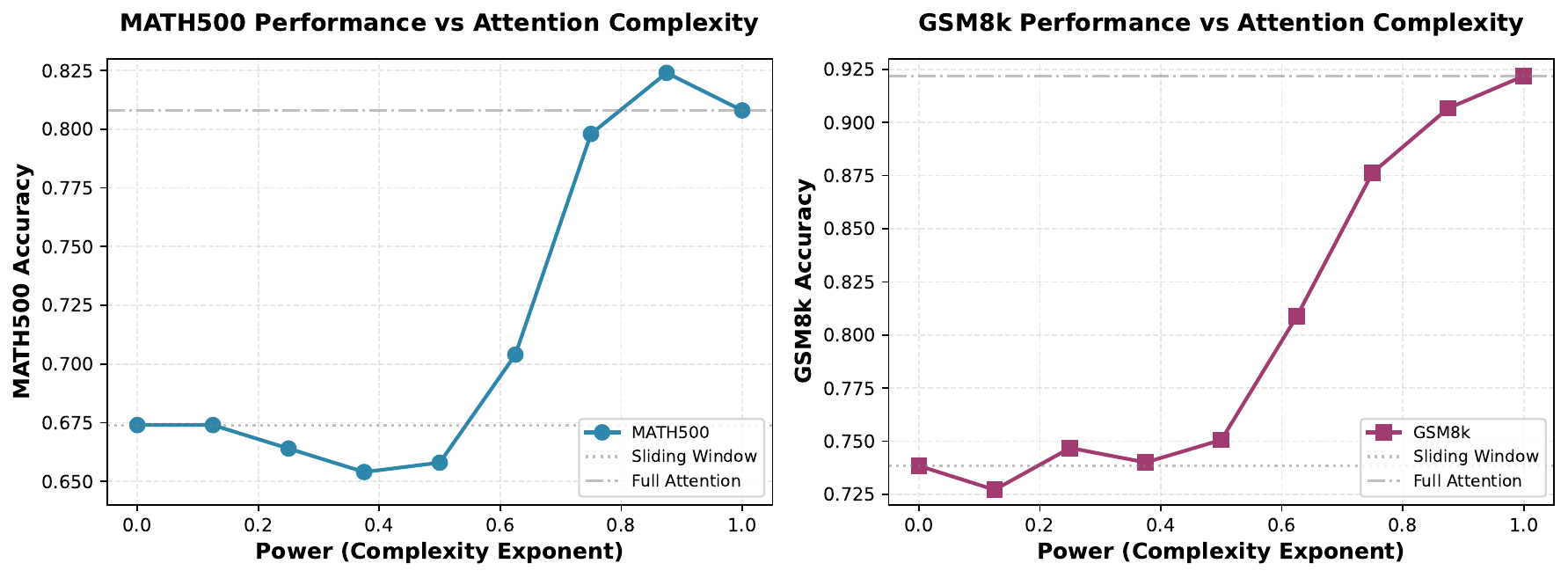}
    \caption{Performance of power-based partial attention (PPA) on MATH500 (left) and GSM8k (right) benchmarks 
    as a function of the power parameter $p$ controlling attention complexity. Both benchmarks exhibit S-shaped curves, 
    with rapid performance gains occurring in the transition region around $p \approx 0.75$--$0.875$. The horizontal reference lines 
    indicate the performance of sliding window attention ($p=0$) and full attention ($p=1$), demonstrating that 
    sub-quadratic attention mechanisms with intermediate $p$ values can achieve near-full-attention performance.}
    \label{fig:ppa-performance}
\end{figure}

After fine-tuning, we evaluated all models on MATH500 and GSM8k benchmarks. Table~\ref{tab:ppa-results} and Figure~\ref{fig:ppa-performance} 
summarize the final performance results.

\begin{table}[htbp]
\centering
\begin{tabular}{l l c c}
\hline
Attention type & Power & MATH500 & GSM8k \\
\hline
Sliding Window $O(L^1)$ & 0     & 0.674 & 0.738 \\
PPA $O(L^{1.125})$      & 0.125 & 0.674 & 0.727 \\
PPA $O(L^{1.25})$       & 0.25  & 0.664 & 0.747 \\
PPA $O(L^{1.375})$      & 0.375 & 0.654 & 0.740 \\
PPA $O(L^{1.5})$        & 0.5   & 0.658 & 0.751 \\
PPA $O(L^{1.625})$      & 0.625 & 0.704 & 0.809 \\
PPA $O(L^{1.75})$       & 0.75  & 0.798 & 0.876 \\
PPA $O(L^{1.875})$      & 0.875 & 0.824 & 0.907 \\
Full Attention $O(L^2)$ & 1     & 0.808 & 0.922 \\
\hline
\end{tabular}
\caption{Performance comparison of power-based partial attention (PPA) variants on MATH500 and GSM8k benchmarks. 
The power parameter $p$ controls attention complexity from $O(L)$ (sliding window, $p=0$) to $O(L^2)$ (full attention, $p=1$). 
Results show that PPA with $p \approx 0.75$--$0.875$ achieves near-full-attention performance while maintaining sub-quadratic complexity.}
\label{tab:ppa-results}
\end{table}

As shown in Table~\ref{tab:ppa-results} and Figure~\ref{fig:ppa-performance}, the overall shape of the performance curve is S-shaped or sigmoid-like. Initially, as we 
increase $p$ from 0, there is little performance gain. At intermediate values of $p$, the performance begins to increase rapidly, approaching the asymptotic
performance of full attention. When $p$ approaches 1, the return diminishes, with marginal gains from further increases in $p$. 

In our sweep, sliding window attention behaves like a local optimum: small increases in $p$ from 0 yield 
no performance improvement. One must increase $p$ by a significant amount to observe meaningful gains in model performance. 
However, even a small increase in $p$ can prevent direct reuse of highly optimized sliding window kernels, 
resulting in substantial computational overhead. 

Conversely, the widely used full attention is not necessarily optimal either, as we observe diminishing returns at large $p$. 
However, it is important to keep in mind that although $O(L^{1+p})$ theoretically scales slower than $O(L^2)$, 
in practice, the highly optimized full attention implementation using FlashAttention is likely 
more efficient than sub-quadratic variants with less-optimized kernels. 

For intermediate values of $p$, these results reveal a rapid transition region (around $p\approx0.75$ in our experiments) where PPA achieves 
near-full-attention performance while remaining sub-quadratic. Practically, this suggests a usable sweet 
spot for $p$ that balances computational and memory costs with accuracy. Future work should quantify latency and memory trade-offs on real 
hardware, validate across more tasks and model sizes, and derive theoretical approximation bounds for PPA. 

While we fine-tuned the model on a prompt format similar to MATH500, the model is under-fitted on the GSM8k-style prompt format.
This is intentional so that we can observe how the optimal value of $p$ depends on training efforts.
As we can see from the result, although the performance trend for the GSM8k task is similar to MATH500, 
the optimal $p$ value appears to be higher. This suggests that insufficient training 
leads to higher optimal $p$ values. Conversely, this also means that additional training can in fact lower the optimal value of $p$. 

This explains why the optimal value of $p$ on our grid is approximately $0.875$ in the MATH500 case. Even for MATH500, we fine-tune the models with only
200k samples. This amount of data is insufficient for the model to fully learn how to leverage the new $O(L^{1+p})$ attention mechanism. More extensive training 
with larger datasets is necessary to lower the optimal value of $p$, potentially making the practical implementation 
competitive with or more efficient than the highly optimized FlashAttention implementation of full attention.  

\section{Conclusion}

We have introduced power-based partial attention (PPA), a novel attention mechanism that enables systematic exploration of the performance-efficiency trade-off in transformer architectures through a single parameter $p$ controlling the attention complexity from $O(L)$ to $O(L^2)$. Our key findings are:

\begin{itemize}
\item Performance transitions from sliding-window (linear-complexity) attention to full attention in an S-shaped curve, with a rapid transition region around $p \approx 0.75$--$0.875$ where near-full-attention performance is achieved while maintaining sub-quadratic complexity.
\item On our grid, sliding window attention ($p=0$) behaves like a local optimum that requires a substantial increase in $p$ to escape, while full attention ($p=1$) exhibits diminishing returns, suggesting neither extreme is necessarily optimal.
\item The optimal value of $p$ depends on training data quantity and prompt structure, with more extensive training potentially lowering the optimal $p$ and improving efficiency.
\end{itemize}

These results challenge the assumption that full quadratic attention is necessary for high performance and suggest that carefully tuned sub-quadratic attention mechanisms can achieve comparable results. However, practical deployment depends on developing optimized kernels for PPA that can compete with highly optimized full attention implementations like FlashAttention. Future work should focus on: (1) developing efficient GPU kernels for PPA, (2) conducting large-scale training experiments to determine the lowest viable $p$ values, (3) validating across diverse tasks and model sizes, and (4) deriving theoretical bounds on the approximation quality of PPA.

% \section*{Acknowledgments}
% Optional acknowledgments section.

% Remove bibliography if not used
\bibliographystyle{plainnat}
\bibliography{references}

\end{document}